\documentclass[runningheads]{llncs}
\usepackage[T1]{fontenc}
\usepackage{graphicx}
\usepackage{latexsym}
\usepackage{amssymb}
\usepackage{amsmath}
\usepackage{booktabs}
\usepackage{enumitem}
\usepackage{graphicx}
\usepackage{color}
\usepackage{xspace}
\usepackage{xcolor}
\usepackage{pifont}
\usepackage{listings}
\usepackage[bottom]{footmisc}
\usepackage{algorithm}
\usepackage{algorithmic}
\usepackage{derivative}
\usepackage{subcaption}
\usepackage{multirow}
\usepackage{makecell}
\usepackage{soul}
\usepackage{relsize}

\newcommand{\BibTeX}{B\kern-.05em{\sc i\kern-.025em b}\kern-.08em\TeX}

\newcommand{\bemph}[1]{\textbf{\emph{#1}}}
\newif\ifcomm

\commtrue
\newcommand{\algo}{\texttt{STV}\xspace}
\newcommand{\algot}{\texttt{STV\textsubscript{T}}\xspace}
\newcommand{\algos}{\texttt{STV\textsubscript{L}}\xspace}

\newcommand{\SHR}[2][]{$\langle #2 \rangle^{#1}$}
\newcommand*\colourcheck[1]{%
  \expandafter\newcommand\csname #1check\endcsname{\textcolor{#1}{\ding{52}}}%
}

\newcommand*\colourcross[1]{%
  \expandafter\newcommand\csname #1cross\endcsname{\textcolor{#1}{\ding{54}}}%
}
\colourcross{red}
\colourcheck{teal}
\usepackage[colorlinks,
citecolor=violet,
urlcolor=violet,
linkcolor=violet,
bookmarks=false,
hypertexnames=true]{hyperref}

\usepackage{color}

\begin{document}
\title{Share Secrets for Privacy: Confidential Forecasting with Vertical Federated Learning}
\titlerunning{Confidential Forecasting with Vertical Federated Learning}
%
%
\author{Anonymous Submission}
\institute{}
\author{Aditya Shankar\inst{1}\orcidID{0009-0009-3046-8724} \and
J\'er\'emie Decouchant\inst{1}\orcidID{0000-0001-9143-3984} \and
Dimitra Gkorou\inst{2} \and
Rihan Hai\inst{1}\orcidID{0000-0002-3720-6585} \and
Lydia Chen\inst{1,3}\orcidID{0000-0002-4228-6735}
} 

%
%
\institute{Delft University of Technology, Delft, The Netherlands\\\email{\{a.shankar,j.decouchant,r.hai\}@tudelft.nl}  \and
ASML, Veldhoven, The Netherlands\\
\email{dimitra.gkorou@asml.com}\\
\and
Université de Neuchâtel, Neuchâtel, Switzerland\\
\email{lydiaychen@ieee.org}}

\maketitle              
\begin{abstract}
Vertical federated learning (VFL) is a promising area for time series forecasting in many applications, such as healthcare and manufacturing. Critical challenges to address include data privacy and over-fitting on small and noisy datasets during both training and inference. Additionally, such forecasting models must scale well with the number of parties while ensuring strong convergence and low-tuning complexity. We address these challenges and propose ``{{S}ecret-shared {T}ime Series Forecasting with {V}FL}'' (STV), a novel framework with the following key features: i) a privacy-preserving algorithm for forecasting with {SARIMAX} and {autoregressive trees} on vertically-partitioned data; ii) decentralised forecasting using {secret sharing} and {multi-party computation}; and iii) novel $N$-party algorithms for matrix multiplication and inverse operations for exact parameter optimization, giving strong convergence with minimal tuning complexity. 
We evaluate on six representative datasets from public and industry-specific contexts. Results demonstrate that STV's forecasting accuracy is comparable to those of centralized approaches. 
Our exact optimization 
outperforms centralized methods, including state-of-the-art {diffusion} models and {long-short-term memory}, by 23.81\% on forecasting accuracy. We also evaluate scalability by examining the communication costs of exact and iterative optimization to navigate the choice between the two. STV's code and supplementary material is available online: \url{https://github.com/adis98/STV}. 

\end{abstract}
\section{Introduction}
\begin{figure}[t]
    \centering
    \includegraphics[width=0.9\textwidth]{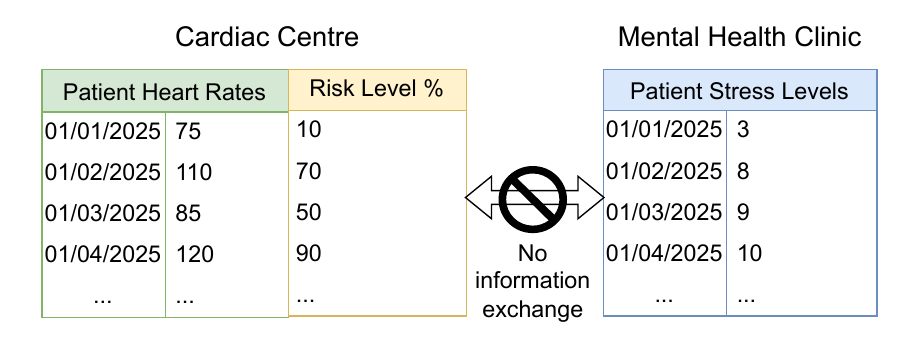}
    \caption{Problem scenario---forecasting risk levels needs corresponding input features from multiple parties, all of whom want to protect the confidentiality of their data.}
    \vspace{3mm}
    \label{fig:problemscenario}
\end{figure}

\textit{Vertically}-partitioned or \textit{feature}-partitioned time series data are prevalent in many areas, such as healthcare, manufacturing, and finance \cite{yang2019federated,tangihvfl,10185757}. For example, consider the scenario in \autoref{fig:problemscenario}, where a mental health facility and a cardiac centre possess distinct features, such as stress levels and heart rate corresponding to a common patient. Each party can forecast patient risk levels using just their own feature sets, but it is likely that combining stress levels and heart rate could lead to better predictive precision due to the correlations between the two~\cite{de2018intriguing}. The straightforward way to use both sources is to first centralize the two feature sets at a server and then train a model. However, privacy restrictions, such as patient confidentiality agreements or laws (e.g. GDPR~\cite{GDPR2016}), prevent data sharing between organizations. Such a scenario naturally extends to other domains, such as joint predictive maintenance in manufacturing\cite{lin2019time,susto2016dealing}, or stock predictions in finance\cite{10185757}.


\emph{Federated learning} (FL)~\cite{konevcny2015federated} is a promising research direction to address the privacy concern. Its training paradigm follows a \textit{model-to-data} approach where data never leaves the premises of the source. Within FL, \emph{Vertical federated learning} (VFL), considers cases where each participant owns different features pertaining to the same sample ID or timestamp~\cite{yang2019federated}, which corresponds to our problem scenario. However, time-series forecasting with VFL has received limited attention~\cite{yan2022multi,tangihvfl} and existing works overlook the following challenges.

First, VFL methods typically employ {deep learning}, which suffers when data is scarce due to overfitting. Data scarcity is a real issue---healthcare and manufacturing environments can be affected by {slow collection} and {noisy measurements}, leading to small datasets~\cite{li2021gaussian,zhu2023co,litjens2017survey}. Second, VFL methods predominantly use a  \emph{split-learning} architecture, consisting of several bottom models at the clients and a single top model held by a server~\cite{vepakomma2018split,hardy2017private,chen2020vafl}. Final predictions are assumed to be generated by the top model at the server. However, since each party may have an interest in maintaining control or ownership over the final output, deciding which party should assume the server role can lead to conflicts. Moreover, centralizing predictions at a single party would be catastrophic in case of a data breach. Third, the models are generally trained iteratively---e.g., via gradient descent---needing extensive hyperparameter tuning. In contrast, analytical/exact methods can reach globally optimal solutions without  any tuning, but may incur higher compute complexity. Therefore, we need a way to flexibly incorporate both approaches to handle diverse scenarios.  

 To address the challenges, we present a novel framework, {\textit{Secret-Shared Time Series Forecasting with Vertical Federated Learning}} (\algo), with the following contributions. \textbf{1. VFL forecasting framework}---built using \emph{secret sharing} (SS)~\cite{shamir1979share} and \emph{multi-party computation} (MPC)~\cite{cramer2015secure,lindell2005secure} for {cryptographic privacy guarantees}. We propose \algos for  linear models such as \textit{SARIMAX}~\cite{Korstanje2021,statsmodels2023,hamilton2020time}, and \algot for \text{autoregressive trees} (ARTs)~\cite{meek2002autoregressive}. \textbf{2. Decentralized inference}---all outputs and intermediate data remain distributed across parties as shares when using MPC. Hence,  responsibility of output generation is also distributed across all the involved parties, promoting trust through mutual dependence. \textbf{3. Adaptable optimization with Least Squares}---using a \textit{two-step} approach that can flexibly be incorporated into both iterative and exact methods. This approach requires algorithms for \textit{$N$-party} \text{matrix multiplications} and \text{inverses} on secret shares, a key novelty in our work.

Our evaluation compares the forecasting accuracy of \algos and \algot with centralized state-of-the-art forecasters based on diffusion models~\cite{alcaraz2022diffusion}, Long Short Term Memory (LSTM), and SARIMAX with Maximum Likelihood Estimation (MLE)~\cite{hamilton2020time}. We also compare the communication costs of iterative and exact optimization of linear forecasters under different scaling scenarios, highlighting their trade-offs. We use a wide range of datasets: five public datasets (Air quality, flight passengers, SML 2010, PV Power, Rossman Sales) and one industrial semiconductor dataset for measuring chip overlays from alignment sensor data.

\section{Background}
In this section, we provide background knowledge on time series models and secure multiparty computation. We provide a summary of our notations and symbols in \autoref{tab:notations}.

\subsection{Time series forecasting}
\label{ssec:SARIMAX}

 \algo uses a popular linear forecaster, SARIMAX ({S}easonal {A}uto{R}egressive {I}ntegrated {M}oving \textbf{A}verage with e{X}ogenous variables) \cite{Korstanje2021,statsmodels2023}, that generalizes other autoregresive forecasters, such as ARMA, ARIMA, and ARIMAX~\cite{hamilton2020time,montgomery2015introduction}. It models the outputs as a linear function composed of {autoregressive}, moving averages, and {exogenous} variables along with their seasonal counterparts~\cite{statsmodels2023}. 
 
 Autoregressive terms capture the influence of historical values on future predictions and moving average terms capture the influence of historical errors/residuals. Exogenous features serve as additional auxiliary features that aid in forecasting. The seasonal terms capture periodic patterns in the series. Mathematically, the output $Y$ at time $t$ can be modelled using a linear polynomial $H$, containing historical residuals $\epsilon(t-i)$, historical observations $Y(t-i)$, and exogenous features $X(t)$ as follows: 

\begin{equation}
\scalebox{0.8}{$
\begin{split}
H: Y(t) {=} \alpha_1Y(t{-}1) {+} \alpha_2Y(t{-}2) {+} \beta_1\varepsilon(t{-}1) {+} \gamma_1X_1(t){+}\gamma_2X_2(t){+}\varepsilon(t) 
\label{eq:sarimax}
\end{split}$}
\end{equation}
The coefficients ($\alpha, \beta, \gamma$) are generally estimated using least-squares~\cite{hannan1984method,lutkepohl2007general,liu2016online,tarsitano2017short} or Maximum Likelihood Estimation (MLE)~\cite{hamilton2020time}. However, implementing MLE with multiparty computation and secret sharing is limited to likelihood functions like the exponential or multivariate normal distributions~\cite{snoke2017providing,lin2010privacy}, since computations with MPC are limited by a small number of supported mathematical operations. However, least-squares optimization is still possible by extending basic MPC protocols such as scalar addition and multiplication (see \hyperref[sec:smpc]{Section 2.2}). First, we transform the datasets, $(X, Y)$, into time-lagged design matrices ($\phi_X,\phi_Y$) representing eq.\eqref{eq:sarimax}: 

\begin{equation}
\scalebox{0.7}{$
\begin{split}
\underbrace{\begin{bmatrix}
    Y(3)\\
    Y(4)\\
    ..\\
    Y(t)
    \end{bmatrix}}_{\displaystyle \phi_Y}= \underbrace{ 
    \begin{bmatrix}
        Y(2) & Y(1) & \mathlarger{\varepsilon(2)} & X_1(3) & X_2(3) \\
        Y(3) & Y(2) & \mathlarger{\varepsilon(3)} &  X_1(2) & X_2(2)\\
        .. & .. & .. & .. & ..\\
        Y(t-1) & Y(t-2) & \mathlarger{\varepsilon(t-1)} &  X_1(t) & X_2(t)
    \end{bmatrix}}_{\displaystyle \phi_X}
     \times \underbrace{\begin{bmatrix}
        \alpha_1\\
        \alpha_2\\
        \beta_1\\
        \gamma_1\\
        \gamma_2\\
    \end{bmatrix}}_{\displaystyle A}+\underbrace{\begin{bmatrix}
        \mathlarger{\varepsilon(3)}\\
        \mathlarger{\varepsilon(4)}\\
        ..\\
        \mathlarger{\varepsilon(t)}
    \end{bmatrix}}_{\displaystyle \mathlarger{\varepsilon}}
    \end{split}$}
    \label{eq:realignment}
\end{equation}
\begin{table}[tb]
    \centering
    \caption{Summary of key mathematical notations}
    \begin{tabular}{|c|c|}
    \hline
       Variable  &  Description\\\hline
        $X, Y$ & Exogenous features and output variable\\
        $H$ & Time series polynomial\\
        $.(t)$ & Variable value at timestep $t$\\
        $\epsilon$ & Residual features\\
        $\alpha, \beta, \gamma$ & Autoregressive, moving average, exogenous coefficients\\
        $\phi_{X}, \phi_{Y}$ & Time-lagged design matrix of features and output\\
        $C_1, C_{i\in[2:K]}$ & Active party, passive parties\\
        \SHR{P} & Secret-shared state of a variable $P$. Union of \SHR[i]{P}\\
        \SHR[i]{P}& $C_i$'s share of a variable $P$\\
        $K$ & Number of parties or clients\\

 \hline

    \end{tabular}
    
    \label{tab:notations}
\end{table}

With least-squares, $A$ can be optimized using a \emph{two-step} regression approach \cite{tarsitano2017short,lutkepohl2007general,hannan1984method,liu2016online}. First, the residuals are estimated by modeling using only autoregressive (AR) and exogenous terms. Then, all the coefficients in $A$ are jointly optimized by setting the residuals to the estimates. 

Since the residuals, $\epsilon(.)$, in $\phi_X$ are unknown, they are initialized to zero to give $\hat{\phi}_X$, as shown below: 

\begin{equation}
\scalebox{0.7}{$
\begin{split}
\underbrace{\begin{bmatrix}
    Y(3)\\
    Y(4)\\
    ..\\
    Y(t)
    \end{bmatrix}}_{\displaystyle \phi_Y}= \underbrace{ 
    \begin{bmatrix}
        Y(2) & Y(1) & 0 & X_1(3) & X_2(3) \\
        Y(3) & Y(2) & 0 &  X_1(2) & X_2(2)\\
        .. & .. & .. & .. & ..\\
        Y(t-1) & Y(t-2) & 0 &  X_1(t) & X_2(t)
    \end{bmatrix}}_{\displaystyle \hat{\phi}_X}
     \times \underbrace{\begin{bmatrix}
        \hat{\alpha_1}\\
        \hat{\alpha_2}\\
        \hat{\beta_1}\\
        \hat{\gamma_1}\\
        \hat{\gamma_2}\\
    \end{bmatrix}}_{\displaystyle \hat{A}}+\underbrace{\begin{bmatrix}
        \mathlarger{\varepsilon(3)}\\
        \mathlarger{\varepsilon(4)}\\
        ..\\
        \mathlarger{\varepsilon(t)}
    \end{bmatrix}}_{\displaystyle \varepsilon}
    \end{split}$}
    \label{eq:twostep}
\end{equation}

With mean-squared-error (MSE), $\hat{A}$ is optimized using the \textit{normal equation} (NE)~\cite{blais2010least} or gradient descent (GD): 

\ul{\emph{Normal equation:}}
\begin{equation}
        \hat{A} = (({\hat{\phi}_X})^{T}{(\hat{\phi}_X)}^{-1}(({\hat{\phi}_X})^{T}{\phi_Y})  
        \label{eq:normal_eqn}
\end{equation}

\ul{\emph{Gradient descent:}}
\begin{equation}
    \hat{A} := \hat{A} - \frac{\alpha}{N} \times (\hat{\phi}_X)^T \times (\hat{\phi}_Y - \phi_Y) \text{  (for $e$ iterations)}
    \label{eq:gdupdate}
\end{equation}
Here, $\hat{\phi_Y} = \hat{\phi}_X\times\hat{A}$ are the predictions at a particular step, $\alpha$ is the learning rate and $N$ is the number of samples.

Following this initial estimate, the residuals, $\varepsilon$, are then obtained as follows:
\begin{equation}
    \varepsilon = \phi_Y - \hat{\phi}_X \times \hat{A}
\end{equation}

These residual estimates are then re-substituted in eq.\eqref{eq:realignment} to refine $\hat{A}$ via a second optimization step. Although we describe the two-step optimization procedure in the context of linear forecasters, the idea of autoregression can also be extended to tree-based models like XGBoost~\cite{chen2015xgboost}.  \textit{Autoregressive trees} (ARTs)~\cite{meek2002autoregressive} build on this premise and use historical outputs/observations as decision nodes in a gradient-boosted regression tree. Therefore, extending  XGBoost-based VFL methods \cite{xie2022efficient,cheng2021secureboost,fang2021large} to ARTs only requires transforming the datasets into lagged design matrices using a polynomial like eq.\eqref{eq:sarimax}. This architecture enables modelling forecasts in a non-linear fashion.


\subsection{MPC}
\label{sec:smpc}
Multi-Party Computation methods use the principle of \textit{secret sharing}~\cite{shamir1979share} for privacy by scattering a value into random shares among parties. These methods offer strong privacy guarantees, i.e., \textbf{information-theoretic security}~\cite{fang2021large}. Assume there are $K$ parties, $C_{i \in [1,K]}$. If party $C_i$ wants to secure its private value, $V$, it does so by generating $K-1$ random shares, denoted \SHR[i']{V}; $\forall i' \in [1,K]; i' \ne i$. These are sent to the corresponding party $C_{i'}$. $C_i$'s own share is computed as \SHR[i]{V} = $V$ - $\sum_{i' \ne i}^{K}$\SHR[i']{V}. The whole ensemble of $K$ shares representing the shared state of $V$, is denoted as \SHR{V}. 

Parties cannot infer others' data from their shares alone, as shares are completely random. However, the value can be recovered by combining all shares. We can extend this idea to machine learning by distributing  feature \textit{vectors} and outputs as shares to preserve their privacy. All parties then jointly utilize decentralized training protocols on the secretly shared data to obtain a local model. Inference/forecasting is then done by distributing features into secret shares and then computing the prediction as a distributed share across all parties. Training and forecasting on secretly shared features requires primitives for performing mathematical operations on distributed shares, which we explain as follows.
 
\bemph{Addition and subtraction}: If $X$ and $Y$ exist as secret shares, \SHR{X} and \SHR{Y}, each party performs a local addition or subtraction, i.e., \SHR[k]{Z} = \SHR[k]{X} $+(-)$ \SHR[k]{Y}. To obtain $Z = X +(-) Y$, the shares are aggregated: $Z = \sum_{k}^{K}$\SHR[k]{Z}. 

    Knowing the value of \SHR[k]{Z} makes it impossible to infer the private values $X$ or $Y$, as each participant only owns a share of the whole secret. Moreover, the individual values of the shares, \SHR[k]{X} and \SHR[k]{Y}, are also masked by adding them.
   
\bemph{Multiplication (using Beaver's triples)}~\cite{beaver1992efficient,xie2022efficient}: Consider $Z=X * Y$, where $*$ denotes element-wise multiplication, and $X$ and $Y$, are secretly shared. The coordinator first generates three numbers $a, b, c$ such that $c = a * b$. These are then secretly shared, i.e., $C_k$, receives \SHR[k]{a}, \SHR[k]{b}, and \SHR[k]{c}. $C_k$ computes \SHR[k]{e} = \SHR[k]{X} - \SHR[k]{a} and \SHR[k]{f} = \SHR[k]{Y} - \SHR[k]{b}, and sends it to $C_1$. $C_1$ then aggregates these shares to recover $e$ and $f$ and broadcasts them to all parties. $C_1$ then computes \SHR[1]{Z} = $e * f$+$f * $\SHR[1]{a}+ $e *$\SHR[1]{b}+\SHR[1]{c}, and the others calculate \SHR[k]{Z} = $f*$\SHR[k]{a}+$e*$\SHR[k]{b}+\SHR[k]{c}. It is easy to see that aggregation of the individual shares gives the product $Z$. Despite knowing $e$ and $f$, $X$ and $Y$ are hidden since the parties do not know $a,b,$ and $c$. Moreover, the values of $X$ and $Y$ are also hidden from the coordinator, who is only responsible for the generation of $a,b$, and $c$, and does not hold any features or outputs. Similar to addition, the individual shares, \SHR[k]{Z}, do not reveal anything about the local share values, i.e., \SHR[k]{X}, \SHR[k]{Y}, \SHR[k]{a}, \SHR[k]{b}, and \SHR[k]{c}. Additional primitives for division and argmax can also be computed using MPC, as detailed in Fang et al.~\cite{fang2021large} and Xie et al.~\cite{xie2022efficient}.

\section{Related Work}

\begin{table}[b]
    \centering
    \caption{Comparison of related works in VFL.}
    \begin{tabular}{c c c c c}
    
        \hfill \rotatebox{10}{Method}
        &\rotatebox{10}{Time Series}& \rotatebox{10}{Serverless inference} & \rotatebox{10}{Dual optimization} & \rotatebox{10}{N-party ($\geq 2$)} \\
        \Xhline{1.5 pt}
        
        Yan et al.~\cite{yan2022multi} &\tealcheck  &\redcross &\redcross &\tealcheck\\

        Han et al.\cite{han2009privacy}& \redcross &
        \tealcheck&\tealcheck &\redcross\\

        Xie et al.\cite{xie2022efficient}&\redcross &\tealcheck &\redcross &\tealcheck\\

        Shi et al.\cite{shi2022mvfls}& \redcross &\tealcheck & \redcross &\tealcheck\\
        
        \hline
        \algo (this work)  &\tealcheck &\tealcheck&\tealcheck &\tealcheck\\
        \Xhline{1.5 pt}
    \end{tabular}
    \label{tab:relwork}
\end{table}

As shown in  \autoref{tab:relwork}, we compare VFL methods on their applicability to time series forecasting, potential for achieving inference privacy, adaptability to alternative optimization choices, and generalizability to $N$-parties.

\bemph{Time-series forecasting.} Earlier, we mentioned that industries require easy-to-understand and low-complexity models to avoid overfitting on small datasets. Hence we focus on linear/logistic regression (LR)~\cite{hardy2017private,han2009privacy,shi2022mvfls}, and tree-based models~\cite{xie2022efficient,cheng2021secureboost}. Yan et al.~\cite{yan2022multi} modifies the split learning architecture using Gated Recurrent Units (GRUs) with a shared upper model for predictions. However, it depends on a single server to produce forecasts, which is a bottleneck and can lead to trust issues among the involved parties. 

\bemph{Decentralized inference.} We consider schemes adopting SS as potential candidates for our inference requirements, as they are straightforward to integrate into a decentralized approach like ours~\cite{xie2022efficient,shi2022mvfls,han2009privacy}. Shi et al.~\cite{shi2022mvfls} and Han et al.~\cite{han2009privacy} train linear models while Xie et al.~\cite{xie2022efficient} implement XGBoost. Homomorphic encryption methods, such as Hardy et al.~\cite{hardy2017private}, require the predictions to be decrypted at the server, violating the requirement.

\bemph{Optimization.} When it comes to optimization techniques, all selected works, except for Han et al.~\cite{han2009privacy}, employ solely iterative approaches. Notably, Han et al.~\cite{han2009privacy} offer iterative and matrix-based methods for exact optimization using eq.\eqref{eq:normal_eqn}. However, they only provide 2-party protocols for matrix multiplications and inverses, which we extend to the $N$-party case.

\section{\algo Framework}
\label{sec:ssvfl}
\begin{figure*}[tb]
    \centering
    \begin{subfigure}{0.4\textwidth}
    \begin{center}
    \includegraphics[width=0.7\textwidth]{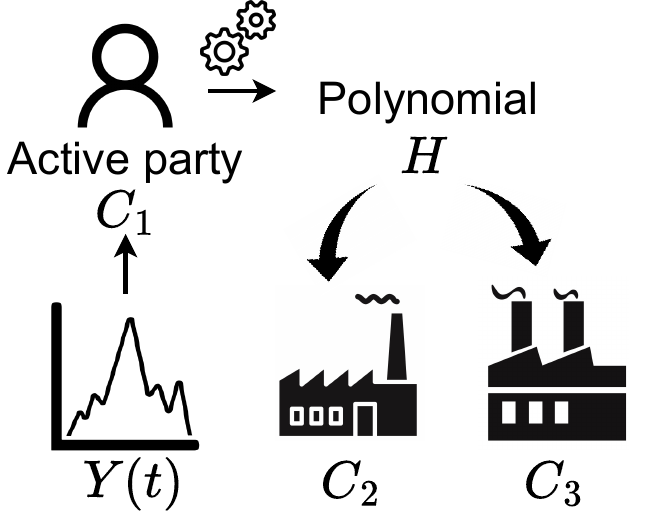}
    \caption{Polynomial generation through pre-processing}
    \label{fig:preprocessing}
        \end{center}
    \end{subfigure}
    \hfill
    \begin{subfigure}{0.4\textwidth}
    \begin{center}
    \includegraphics[width=1.0\textwidth]{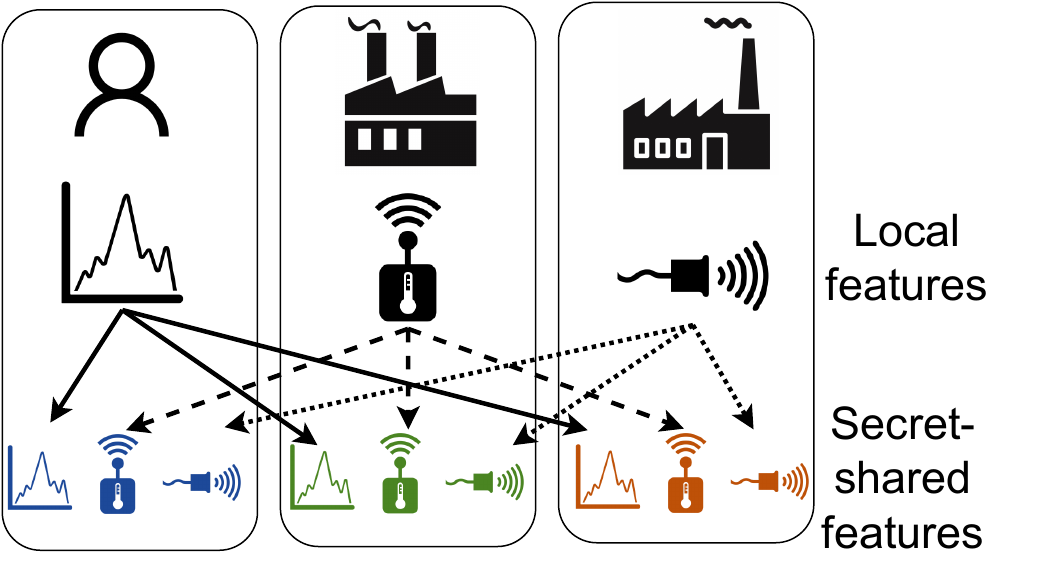}
    \caption{Secretly sharing features}
    \label{fig:secretsharing}
        \end{center}
    \end{subfigure}
\caption{Time-series pre-processing and secret sharing of features in \algo}
\label{fig:stv}
\end{figure*}
\begin{figure*}
\centering
\begin{subfigure}{0.4\textwidth}
\begin{center}
\includegraphics[width=1.3\textwidth]{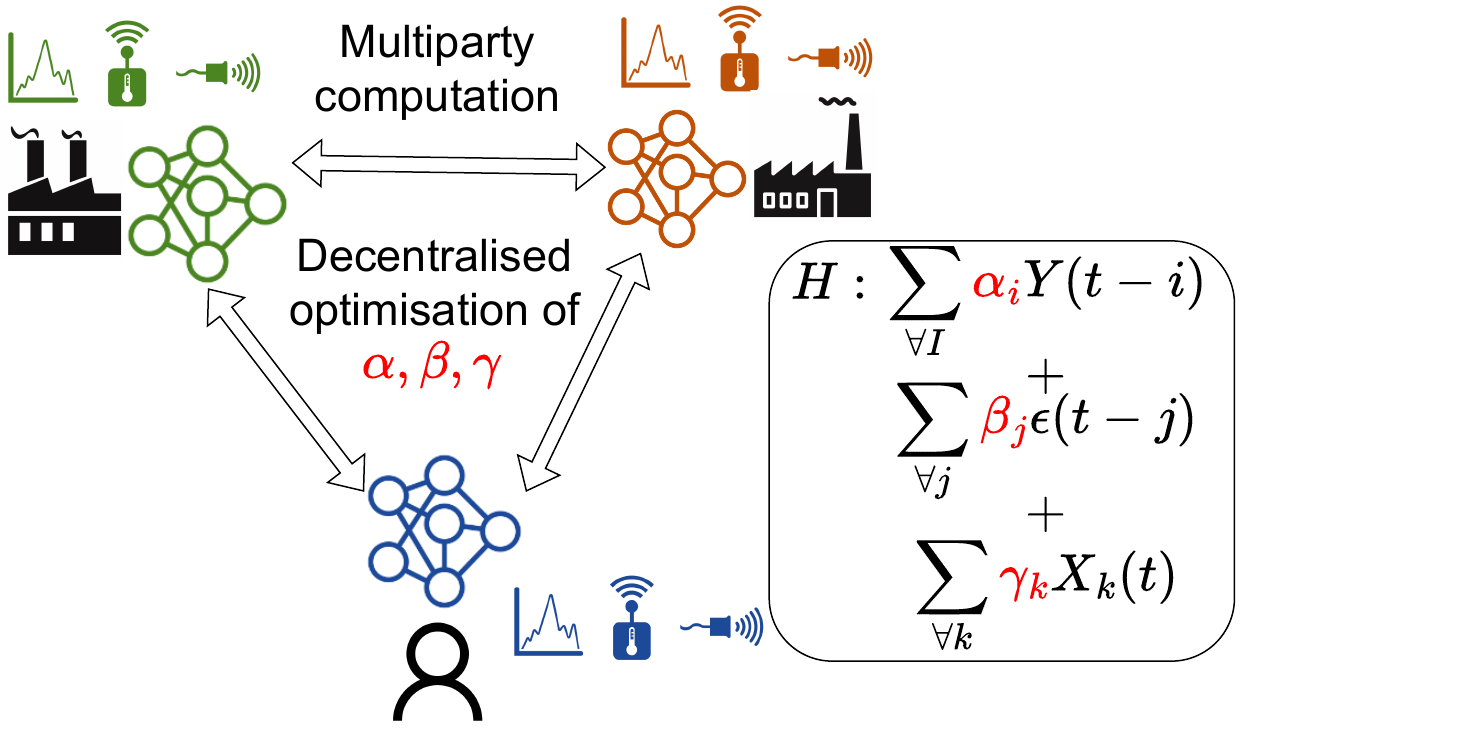}
\caption{MPC optimization of time-series polynomial $H$ with secret-shared features}
\label{fig:training}
\end{center}
\end{subfigure}
\hfill
\begin{subfigure}{0.4\textwidth}
\begin{center}
    \includegraphics[width=0.9\textwidth]{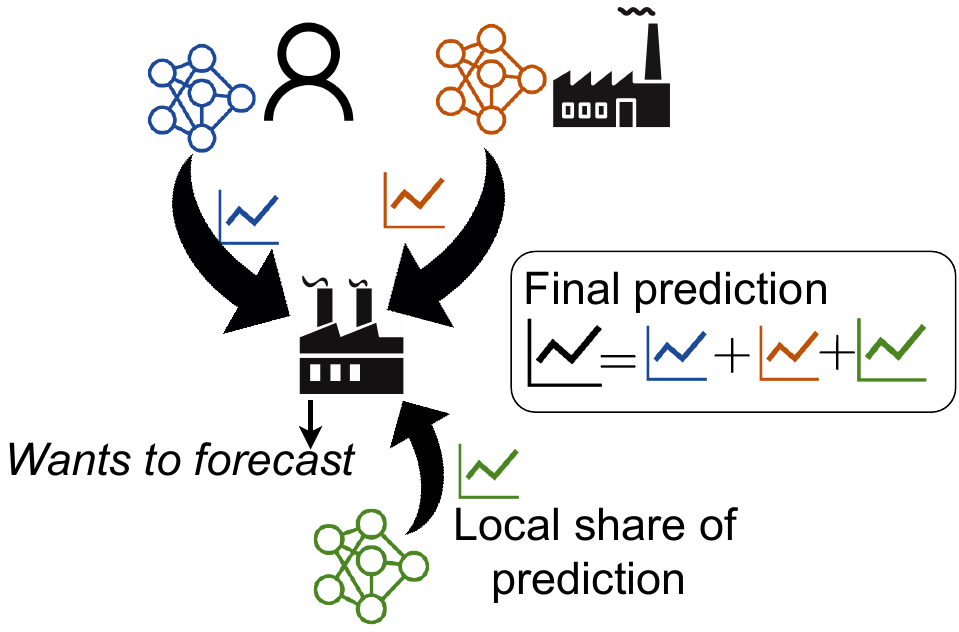}
    \caption{Decentralized inference through flexible selection of share-aggregator}
    \label{fig:inference}
\end{center}
\end{subfigure}
\caption{Training and inference in \algo}
\label{fig:trainandinfer}
\end{figure*}
\begin{algorithm}[tb]  
\caption{General protocol \algo}
    
    \label{alg:alg1}
    \begin{flushleft}
    \textbf{Data}: $X_k$ on party $C_k$ $\forall k \in [1,K]$, and $Y$ on $C_1$\\
    \textbf{Accepted Parameters}: Task: \emph{Training/Inference}, Model \emph{type}, number of trees $T$,
    Optimization method $O$,
    learning rate $\alpha$, iterations, $e$, Trained distributed $Model$, Requesting party $C_j$ \\
    \textbf{Output}:Trained  model distributed across $K$ parties or predicted forecast at  party $C_j$\\
    \end{flushleft}
    \begin{algorithmic}[1] 
    \IF {$Training$ and active party $C_1$}
    \STATE \emph{ params = ProcessSeries(Y)} \label{ln:procseries}
    \STATE \emph{H = GenPoly (params, type)} \label{ln:genpoly}
    \STATE Broadcast $H$ to $C_i$ $\forall i \in [1,K]$ 
    \ENDIF
    \STATE Share local features \SHR{X_k} and (or) outputs \SHR{Y} 
    \STATE \SHR{\phi_X},\SHR{\phi_Y} = \emph{TransformData} ($H$)\COMMENT{ \eqref{eq:realignment}} \label{ln:transform}
    \IF {\emph{type} == \emph{Tree} and \emph{Training}}
    \RETURN $Model$ = \emph{\algot(\SHR{\phi_X}, \SHR{Y},T)}
    \ELSIF{\emph{type} == \emph{Linear} and \emph{Training}}
    \RETURN $Model$ = \emph{\algos(\SHR{\phi_X},\SHR{\phi_Y}, O, $\alpha$, $e$)}
    \ENDIF
    \IF {\emph{Inference}}
    \STATE \SHR{Result} = $Model.Predict$(\SHR{\phi_X}) \label{ln:localpredict}
    
    \IF{requesting party $C_j$}
    \STATE $Result$ = $\sum_{k=1}^K$\SHR[k]{Result} \COMMENT{Aggregate predictions} \label{ln:aggpredict}
    \ENDIF
    \ENDIF
    \end{algorithmic}
    \end{algorithm}
\begin{algorithm}[tb]
    \caption{\algot}
    \label{alg:alg2}
    \begin{flushleft}
    \textbf{Data}: Secretly shared transformed matrices \SHR{\phi_X}, \SHR{\phi_Y}  \\
    \textbf{Parameter}: number of trees $T$\\
    \textbf{Output}: Distributed autoregressive XGBoost tree \\
    \end{flushleft}
    \begin{algorithmic}[1] 
    
    \STATE Initialize predictions \SHR[k]{\hat{\phi}_Y}$=0$ on all parties $C_k$
    \STATE Initialize {$Trees_k$} = [ ] on all parties $C_k$
    \FOR{$t \in [1,T]$}
    \STATE $tree_{t_k}$ = \bemph{SecureFit(\SHR[k]{\phi_X},\SHR[k]{\phi_Y},\SHR[k]{\hat{\phi}_Y})} \label{ln:securefit}
    \STATE \SHR[k]{\hat{\phi}_{Y_t}} = $tree_{t_k}.$\emph{Predict(\SHR[k]{\phi_X})} \label{ln:treepredictlocal}
    \IF{active party $C_1$}
    \STATE $\hat{\phi}_{Y_t} = \sum_{i=1}^{K}$\SHR[k]{\hat{\phi}_{Y_t}}\COMMENT{Aggregate predictions} \label{ln:treeaggpreds}
    \STATE $\hat{\phi}_{Y} = \hat{\phi}_{Y} + \hat{\phi}_{Y_t}$\COMMENT{Add to final predictions}
    \ENDIF
    \STATE $Trees_k.append(tree_{t_k})$ on every party $C_k$
    \ENDFOR
    \RETURN $Trees_k$ on party $C_k$

    \end{algorithmic}
    \end{algorithm}
\begin{algorithm}[tb]
    \caption{\algos}
    \label{alg:alg3}
    \begin{flushleft}
    \textbf{Data}: Secretly shared transformed matrices \SHR{\phi_X}, \SHR{\phi_Y}
    \\
    \textbf{Accepted Parameters}:  $O$, $\alpha$, $e$ \\
    \textbf{Output}: Shared optimized coefficients \SHR{A}\\
    \end{flushleft}
    \begin{algorithmic}[1] 
    \FOR{$step \in [1,2]$}
    \IF{$step = 1$}
    \STATE Initialize residuals to zero in \SHR[k]{\phi_X} for all $C_k$
    \ENDIF
    \IF {$O=``iterative"$}
    \STATE Randomly initialize \SHR[k]{A} for all $C_k$ \label{ln:rndminit}
    \FOR{$e$ iterations}
    \STATE Get \SHR{\hat{\phi_Y}} = \SHR{\phi_X}$\times$\SHR{A} using Alg.~\ref{alg:alg4} \label{ln:prediter}
    \STATE \SHR{\odv{l}{A}} = (\SHR{{\phi_X}})$^T$$\times$$($\SHR{\hat{\phi_Y}} - \SHR{\phi_Y}$)$ (Alg.~\ref{alg:alg4}) \label{ln:graditer}
    \STATE Perform update: \SHR{A} := \SHR{A} - $\frac{\alpha}{N}$\SHR{\odv{l}{A}} \label{ln:updateiter}
    \ENDFOR
    \ELSIF{$O=``exact"$}
    \STATE \SHR{Z} = (\SHR{\phi_X})$^T\times$\SHR{\phi_X} using Alg.~\ref{alg:alg4} \label{ln:transposedir}
    \STATE  \SHR{W} = \SHR{Z^{-1}} using Alg.~\ref{alg:alg5} \label{ln:invdir}
    \STATE  \SHR{V} = (\SHR{\phi_X})$^T\times$\SHR{\phi_Y} using Alg.~\ref{alg:alg4} \label{ln:vdir}
    \STATE  \SHR{A} = \SHR{W} $\times$ \SHR{V} \label{ln:adir}
    \ENDIF
    \IF{$step = 1$}
    \STATE Predict: \SHR{\hat{\phi_Y}} = \SHR{\phi_X} $\times$ \SHR{A} using Alg.~\ref{alg:alg4} \label{ln:preddir}
    \STATE Estimate residuals \SHR{\varepsilon} = \SHR{\phi_Y} - \SHR{\hat{\phi_Y}}
    \STATE Set \SHR[k]{\phi_X} using residuals from \SHR[k]{\varepsilon} for all $C_k$
    \ENDIF
    \ENDFOR
    \RETURN \SHR[k]{A} on all parties $C_k$
    \end{algorithmic}
    \end{algorithm}
\begin{algorithm}[bt]
    \caption{Secure Matrix Multiplication}
    \label{alg:alg4}
    \begin{flushleft}
    \textbf{Data}: Secretly shared matrices \SHR{U} and \SHR{V} across $K$ parties, $C_1,C_2,..,C_K$\\
    \textbf{Output}: \SHR{W}, i.e., product $W=U\times V$ as shares across $K$ parties \\
    \end{flushleft}
    \begin{algorithmic}[1] 
    \FOR{each row index $i$}
    \FOR{each column index $j$}
    \STATE \SHR{\vec{T}} = \SHR{\vec{U[i,:]}} * \SHR{\vec{V[:,j]}} \COMMENT{Element-wise product} \label{ln:matrixmul}
    \SHR[k]{W_{i,j}} = $sum\{$\SHR[k]{\vec{T}}$\}$ 
    \ENDFOR
    \ENDFOR
    \RETURN \SHR[k]{W} on party $C_k$
    \end{algorithmic}
    \end{algorithm}
\begin{algorithm}[bt]
    \caption{Secure Matrix Inverse}
    \label{alg:alg5}
    \begin{flushleft}
    \textbf{Data}: Secretly shared matrix \SHR{U} across $K$ parties, $C_1,C_2,..,C_K$\\
    \textbf{Output}: \SHR{V}, i.e., inverse, $V=U^{-1}$, as a distributed share\\
    \end{flushleft}
    \begin{algorithmic}[1] 
    \IF{active party ($P_1$)}
    \STATE Generate random non-singular perturbation matrix $P$ and secretly share as \SHR{P} \label{ln:perturb}
    \ENDIF
    \STATE Get \SHR{Q} = \SHR{U} $\times$ \SHR{P} using Alg.~\ref{alg:alg4} \label{ln:up}
    \STATE       Aggregate $Q$ = $\sum_{k=1}^{K}$\SHR[k]{Q} on passive party $C_j;j > 1$ \label{ln:aggup}
    \IF{Passive party $C_j$}
    \STATE Compute $R=Q^{-1}=(UP)^{-1} = P^{-1}U^{-1}$ \label{ln:upinv}
    \STATE Generate shares \SHR{R} \label{ln:shrupinv}
    \ENDIF
    
    \STATE Compute \SHR{T} = \SHR{U^{-1}} = \SHR{P} $\times$ \SHR{R} using Alg.~\ref{alg:alg4} \label{ln:res}
    \RETURN \SHR[k]{T} on party $C_k$
    
    \end{algorithmic}
    \end{algorithm}

Here we introduce the adversarial model and problem statement, followed by \algo's design and implementation for trees (\algot), and SARIMAX (\algos). Both models employ secret sharing and MPC to protect the privacy of features. 

\textbf{Adversarial Models.}
 We assume that all parties are \emph{honest-but-curious/semi-honest}~\cite{hardy2017private,yang2019federated}, i.e., they \text{adhere to protocol} but  try to infer others' private data using their own local data and whatever is communicated to them. Also, it is assumed that parties \text{do not collude}. This is a standard assumption in VFL as all parties are incentivised to collaborate due to their mutual dependence on one another for training and inference~\cite{yang2019federated,hardy2017private}. In addition, we also assume that communication between parties is encrypted to prevent snooping.

\textbf{Problem Statement.}
We assume a setup with $K$ parties, $C_1$ to $C_K$, grouped into two types: $active$ and $passive$. Without loss of generality, we denote the active party as $C_1$, which holds the ground truth outputs of the training data for the time series, $Y(t)$, for a timestep $t$. It also possesses exogenous features, $X_1(t)$. The passive parties only have exogenous features, $X_i, \forall i \in [2,K]$. The common samples between parties are assumed to be already identified using privacy-preserving entity alignment approaches~\cite{hardy2017private,scannapieco2007privacy}. Our goal is to forecast future values using exogenous and autoregressive features without sharing them with others in plaintext. We further assume that there is a \emph{coordinator}, a trusted third party that oversees the training process and is responsible for generating randomness, such as Beaver's triples for element-wise multiplication with MPC~\cite{beaver1992efficient,fang2021large}. However, this coordinator cannot access private data and intermediate results, so it does not pose a privacy threat, as mentioned in Xie et al.~\cite{xie2022efficient}. 

\subsection{Protocol Overview}

An overview of \algo is provided in \hyperref[alg:alg1]{Algorithm 1}. The framework consists of preliminary steps for {pre-processing} the series and \text{secretly sharing} features (\autoref{fig:stv}), followed by distributed \textbf{training} and \textbf{inference} (\autoref{fig:trainandinfer}).

\bemph{Training. }The active party initiates by pre-processing the output and determines parameters like the auto-correlations and partial auto-correlations of the series (line \ref{ln:procseries}) to identify the time lags of the autoregressive and moving average terms in eq.\eqref{eq:sarimax}. As shown in \autoref{fig:preprocessing}, these are then used to generate a polynomial for SARIMAX or ARTs (line \ref{ln:genpoly}). As illustrated in \autoref{fig:secretsharing}, the features and outputs are then secretly shared and transformed into lagged design matrices (line \ref{ln:transform}), like in eq.\eqref{eq:realignment}. All parties then follow decentralized training protocols, \hyperref[alg:alg2]{Algorithm 2} or \hyperref[alg:alg3]{Algorithm 3} to train a distributed model. Training is illustrated in \autoref{fig:training}.

\bemph{Privacy-preserving inference. }During inference (\autoref{fig:inference}), the final prediction exists as a distributed share (line \ref{ln:localpredict}), which is aggregated on the requesting party (line \ref{ln:aggpredict}). Since the result is computed as a distributed share, the outputs are not tied to a particular party. 
 
    \subsection{\algot: Autoregressive tree (ART)}
  \algot focuses on the autoregressive tree and 
  transforms the original datasets, $X$ and $Y$, into time-lagged design matrices. Following the training framework in Xie et al. \cite{xie2022efficient}, we train a distributed autoregressive XGBoost model using the secretly-shared design matrices. The details are provided in \hyperref[alg:alg2]{Algorithm 2}.
Training proceeds iteratively, finally resulting in the generation of $T$ trees on each party. At each step, every party learns a new tree and makes a local prediction (lines \ref{ln:securefit}-\ref{ln:treepredictlocal}). The active participant aggregates the distributed shares of the prediction to compute the first and second order gradients needed for generating the next distributed tree. These gradients are then secretly shared with the other parties for them update their local tree ensembles. Details on the tree-building function, \bemph{SecureFit}, are provided in \hyperref[xgboostappendix]{Appendix C}.
    During training, individual predictions are aggregated on the active party since gradients are computed by $C_1$ (see \hyperref[xgboostappendix]{Appendix C}). For inference, aggregation can be performed on any party since gradients are not calculated.

\subsection{\algos: SARIMAX }
\label{subsec:ssvfll}

With \algos, the objective is optimizing the coefficients, $A$, in \autoref{eq:realignment}. This can be done either analytically (exactly) or iteratively using the two-step regression process explained in \hyperref[ssec:SARIMAX]{Section 2.1}. Details are provided in \hyperref[alg:alg3]{Algorithm 3}. 

\hyperref[alg:alg3]{Algorithm 3} requires matrix operations like \textit{multiplication} (lines \ref{ln:prediter}-\ref{ln:graditer}; \ref{ln:transposedir}-\ref{ln:preddir}) and \textit{inverse} (line \ref{ln:invdir}) on secretly shared data. This is a requirement for both iterative and exact methods, as seen in \autoref{eq:normal_eqn} and \autoref{eq:gdupdate}. The \textit{$N$}-party ($\ge 2$) algorithms for performing these operations are detailed below.

\bemph{N-party matrix multiplication.}
To perform matrix multiplication with secretly shared data (algorithm \ref{alg:alg4}), we view the computation of every output element $W_{i,j}$ as a scalar product of row and column vectors (line \ref{ln:matrixmul}). This is
implemented using the $N$-party element-wise product of the row and column vectors with Beaver's triples followed by an addition, as shown in \hyperref[sec:smpc]{Section 2.2}.

\bemph{N-party matrix inverse.} As shown in \hyperref[alg:alg5]{Algorithm 5}, we compute the inverse of a secretly-shared matrix, $U$, using a non-singular perturbation matrix, $P$, generated by the active party (line \ref{ln:perturb}). Subsequently, the aggregation of product $UP$ on a passive party does not leak $U$ as $P$ is unknown (line \ref{ln:aggup}). $(UP)^{-1}$ can then be computed locally and secretly shared, followed by a matrix multiplication with $P$, i.e., $P\times(UP)^{-1} = U^{-1}$, giving the result as a secret share (lines \ref{ln:upinv}-\ref{ln:res}).

Similar to the tree-based models, forecasting is done by secretly sharing features and computing the output as a distributed share (see \autoref{fig:inference}). The true value of the forecast can be obtained by summing the shares across all parties, avoiding the need for having a server act as a middleman for producing forecasts.

\section{Performance Evaluation}
We evaluate the forecasting accuracy of \algo against centralized approaches. We also compare the scalability of iterative and exact optimization for linear forecasters using the total communication cost.
\begin{figure*}[!htb]
\centering
    \begin{subfigure}[h]{0.49\textwidth}
        \centering
        \includegraphics[width=\textwidth]{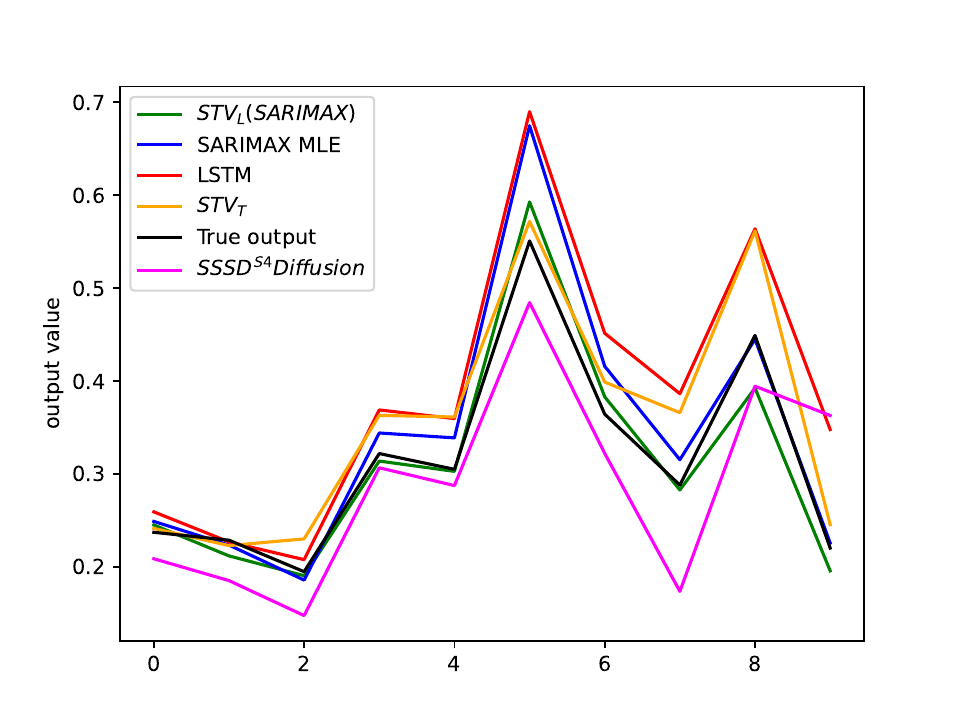}
        \caption{Airquality (50)}
       \label{subfig:airq50}\end{subfigure} 
    \begin{subfigure}[h]{0.49\textwidth}
    \centering
    \includegraphics[width=\textwidth]{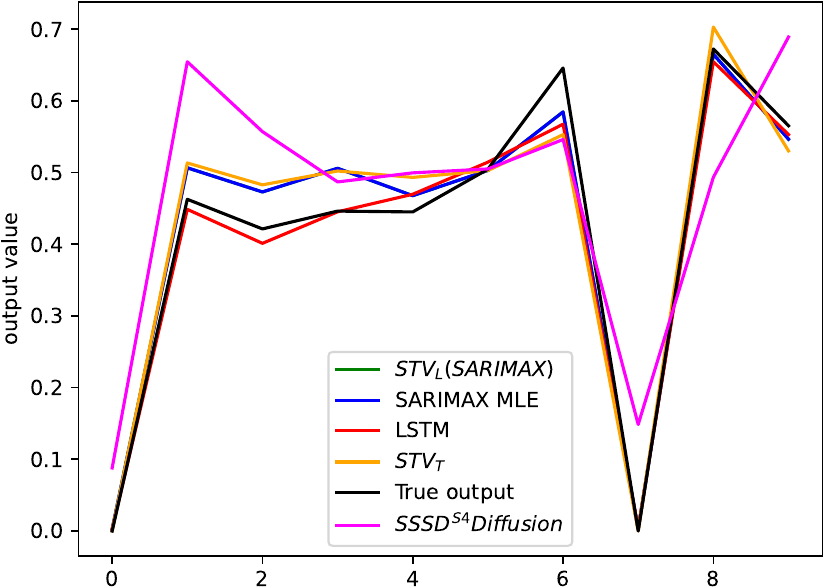}
    \caption{Rossman Sales (50)}
    \label{subfig:ross50}
    \end{subfigure}
    \begin{subfigure}[h]{0.49\textwidth}
    \centering
    \includegraphics[width=\textwidth]{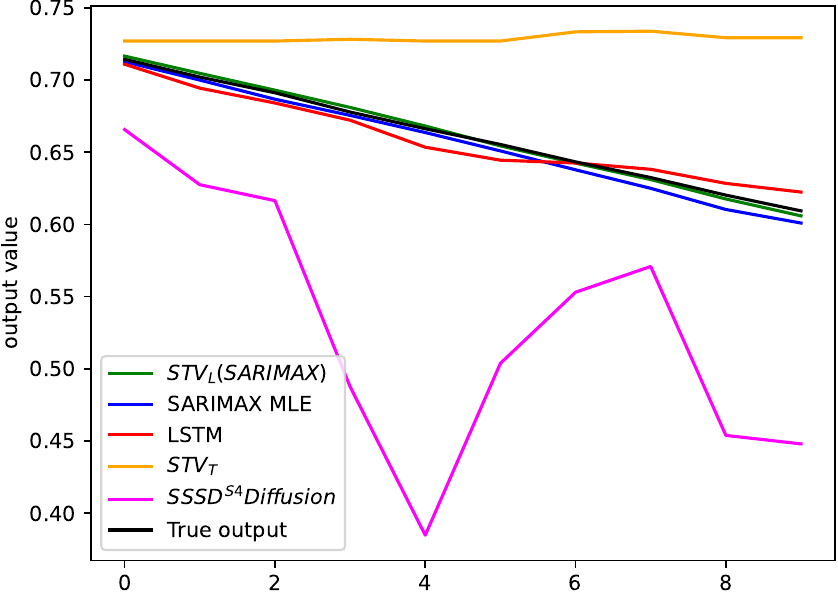}
    \caption{SML 2010 (50)}
    \label{subfig:sml50}
    \end{subfigure}
    \vspace{3mm}
\caption{Forecasts (y-axis) of different methods, total window size 50 (40 steps horizon and forecasts 10 steps ahead). The x-axis shows the forecasted timesteps.}
\label{fig:msefig}   
\end{figure*}

\begin{table*}[!bt]
\caption{Average $n$-MSE and (standard deviation) results for different datasets and methods. Relative improvement of the best VFL method with the best centralized one is also shown (rel. imp). Lowest MSE values are highlighted in bold. SML 2010, Air quality, and Rossman Sales have prequential window sizes 50, 100, 200, and 400. PV Power uses 25, 50, 100, and 200. Airline passengers uses 60, 80, 100, 120, and 140. Finally, the industrial data uses 25, 50, and 100.}
\vspace{3mm}
    \footnotesize
    \centering
    \scalebox{0.75}{
    \begin{tabular}{c c c c c c c}
         \Xhline{1.5pt}
         \multirow{2}{*}{\emph{Dataset}} & \emph{\algos}  & \emph{\algot} & \emph{SARIMAX}  & \emph{LSTM} & \emph{SSSD\textsuperscript{S4}} & Rel. imp. (\%) \\
         & (SARIMAX, VFL) & (ART, VFL) & (MLE, Centralized) & (Centralized) & (Centralized) &\\
         \cline{1-7}
         \multirow{2}{*}{Airquality} & \textbf{0.00069}	&0.00100	&0.00088	&0.00114	&0.00150
         &  \textcolor{teal}{21.59}
\\
         &(0.00008)	&(0.00056)	&(0.00031)	&(0.00076)	&(0.00073) & 
\\\cline{2-7}
         \multirow{2}{*}{Airline passengers} & 0.00304	&0.00808	&\textbf{0.00222}	&0.04130	&0.00392 &  \textcolor{red}{-36.94}
\\
         & (0.00086)	&(0.00379)	&(0.00148)	&(0.04086
)	&(0.00226) & \\\cline{2-7}

         \multirow{2}{*}{PV Power} &\textbf{0.00138} &0.00249 &0.00159 &0.14333 &0.00167
         &  \textcolor{teal}{15.22}
\\
         & (0.00136)	&(0.00254)	&(0.00095)	&(0.23033)	&(0.00058) & \\\cline{2-7}

         \multirow{2}{*}{SML 2010} &\textbf{0.00787}	&0.01875	&0.01033	&0.01716	&0.01085
         &  \textcolor{teal}{23.81}
\\
         &(0.00711)	&(0.01458)	&(0.01061)	&(0.01086)	&(0.00871) &
\\\cline{2-7}
         \multirow{2}{*}{Rossman Sales} 
         &\textbf{0.00074}	&0.00243	&0.00077	&0.00639	&0.00331
         &  \textcolor{teal}{3.89}
\\
         &(0.00012)	&(0.00153)	&(0.00006)	&(0.00280)	&(0.00151) & 
\\\cline{2-7}
         \multirow{2}{*}{Industrial data} 
         &\textbf{0.00602}	&0.04118	&0.00969	&0.00617	&0.01875
         & \textcolor{teal}{2.43}
\\
         &(0.00049)	&(0.04051)	&(0.00449)	&(0.00203)	&(0.00313)   
         &
\\\Xhline{1.5pt}

    \end{tabular}}
    
    \label{tab:msetab}
\end{table*}

\subsection{Forecasting accuracy}
We compare the performance of \algos (exact optimization) and \algot with other centralized methods: Long-Short-Term Memory (LSTM), SARIMAX with MLE\footnote{https://www.statsmodels.org/devel/generated/statsmodels.tsa\\.statespace.sarimax.SARIMAX.html}, and diffusion models for forecasting (\emph{SSSD\textsuperscript{S4}})~\cite{alcaraz2022diffusion}.
The LSTM has two layers with 64 units, followed by two dense layers of sizes 32 and 1. We train the model for 500 epochs with a batch size of 32 and a learning rate of 0.001. The diffusion configuration has the following settings\footnote{Using https://github.com/AI4HealthUOL/SSSD.git}: $T=200, \beta_0=0.0001, \beta_T=0.02$. The model uses two residual layers, four residual and skip channels, with three diffusion embedding layers of dimensions $8 \times 16$. Training is done for 4000 iterations with a learning rate of 0.002.

Five public forecasting datasets are used: Airline passengers~\cite{airlinepassengers}, Air quality data~\cite{misc_air_quality_360}, PV Power~\cite{pvpower}, SML 2010~\cite{misc_sml2010_274}, and Rossman Sales~\cite{rossmansales}. An industry-specific dataset to estimate a performance parameter for \text{chip overlays} from inline sensor values in \text{semiconductor manufacturing} is also included~\cite{gkorou1,gkorou2020get}. Evaluation and pre-processing steps are given in \hyperref[dataappendix]{Appendix A} and \hyperref[experimentappendix]{Appendix B}.
A variant of prequential window testing~\cite{cerqueira2020evaluating} is used since industrial time series data can significantly change after intervals due to machine changes/repairs. The dataset is partitioned into multiple windows of a given size, each further divided into an 80-20 train-test split. After forecasting the test split, the model is retrained on the next window (cf. \hyperref[experimentappendix]{Appendix B}).
All features and outputs are scaled between 0 and 1 for a consistent comparison. We thus present the predictions' normalized mean-squared errors ($n$-MSE). Ground truths on the test set are measured and averaged across multiple windows. We average the $n$-MSE scores across different window sizes to generalize performance scores across varying forecasting ranges, as shown in \autoref{tab:msetab}.  

\autoref{tab:msetab} demonstrates that our approach \algos outperforms other centralized methods in terms of performance. We use exact optimization since iterative gradient descent eventually converges to the same value in the long run. Due to its guaranteed convergence, exact optimization improves against centralized methods by up to 23.81\%. Though it does worse than SARIMAX MLE on the passenger data, it is still among the top two methods. 
Regression plots from a prequential window of size 50 from three datasets are shown in \autoref{fig:msefig}. Generally, all methods can capture patterns in the time series, like in \autoref{subfig:airq50} and \autoref{subfig:ross50}. Occasionally, deep models such as \emph{SSSD\textsuperscript{S4}} overfit, showing the drawback of NNs on small windows (see \autoref{subfig:sml50}).

\subsection{Scalability}

We measure the total communication costs of exact optimization using the normal equation (NE) and iterative batch gradient descent (GD) to analyze the scalability of the two methods under increasing parties, features, and samples. 
We vary the parties between 2, 4, and 8, features between 10 and 100, and samples between 10, 100, and 1000. Since communication costs depend only on the dataset dimensions and the number of parties,  we generate random data matrices for all valid combinations of features and samples on each party, i.e., \#features $\leq$ \#samples. We then optimize the coefficients using either the exact approach (\hyperref[alg:alg3]{Algorithm 3}, lines (\ref{ln:transposedir}-\ref{ln:preddir})), or batch gradient descent (lines (\ref{ln:rndminit}-\ref{ln:updateiter})), for a different number of iterations (10, 100, 1000). For party-scaling, we average the total communication costs across various feature and sample combinations for a given number of parties. Similarly, we measure feature and sample scaling by averaging the total costs across different (sample, party) and (feature, party) combinations. Results are shown in 
\autoref{tab:featurescaling}.

\begin{table}[t]
\caption{Network overhead (B) with varying parties, feature counts and samples.}
\vspace{3mm}
    \footnotesize
    \centering
    \scalebox{0.9}{
    \begin{tabular}{|c|c|c|c|c|c|}

    \cline{3-6}
         \multicolumn{1}{c}{} & &\multirow{2}{*}{NE} & \multicolumn{3}{c|}{GD iterations}  \\\cline{4-6}
         \multicolumn{1}{c}{} & & & 10 & 100 & 1000\\ \cline{1-6}
         \multirow{3}{*}{Parties}  & 2 &	2.54$\times 10^8$	&2.33$\times 10^7$ &2.33$\times 10^8$ &2.33$\times 10^9$ \\ 
         & 4	&5.85$\times 10^8$	&4.65$\times 10^7$ &4.65$\times 10^8$ &4.65$\times 10^9$ \\
         & 8	&1.48$\times 10^9$
         &9.31$\times 10^7$ &9.31$\times 10^8$	&9.31$\times 10^9$ \\\hline

    \hline
         \multirow{2}{*}{Features} & 10	&9.77$\times 10^6$	&8.34$\times 10^6$ 	&8.34$\times 10^7$ 	&8.34$\times 10^8$ 
\\ 
         & 100	&1.92$\times 10^9$	&1.23$\times 10^8$ &1.23$\times 10^9$	&1.23$\times 10^{10}$
\\ \cline{2-6}

    \hline
         \multirow{3}{*}{Samples} & 10	&1.16$\times 10^6$	&2.76$\times 10^5$ &2.76$\times 10^6$	&2.76$\times 10^7$ 
\\ 
         & 100	&4.53$\times 10^8$	&1.24$\times 10^7$	&1.24$\times 10^8$	&1.24$\times 10^9$ 
\\ 
         & 1000	&1.48$\times 10^9$	&1.23$\times 10^8$	&1.23$\times 10^9$	&1.23$\times 10^{10}$ \\\hline
    \end{tabular}}
    
    \label{tab:featurescaling}
\end{table}

    


We observe that when the number of parties/samples/features is small, exact optimization's cost is comparable to an iterative version with more iterations. For example, when the number of parties is 2, we see that the cost of NE is close to GD with 100 iterations.
However, when we increase to 8 parties, the cost of NE exceeds GD with 100 iterations.
Similarly, 
we see that NE has a lower cost than GD with 100 iterations for smaller magnitudes of features and samples. This is no longer the case when increasing the features and samples. 
However, the cost of GD increases proportionally with the iterations, eventually surpassing NE, as seen with 1000 iterations. In practice, hyperparameters like the learning rate and batch size affect the convergence rate, which may be hard to tune in a distributed setup. So, choosing between iterative or exact optimization depends on several factors, requiring adaptability in frameworks.

\section{Conclusion}

We present a generalizable, novel decentralized and privacy-preserving forecasting framework. For instance, healthcare centers can share patient data to assess risks, but confidentiality agreements limit data sharing.
Our results show that the approach competes well with centralized methods. Scalability analyses reveal the trade-offs between iterative and exact optimization, emphasizing the need for adaptability.
Future directions include enabling deep models, such as LSTMs and diffusion, in VFL and exploring hybrid approaches that combine horizontal and vertical FL for improved forecasting. Additionally, moving beyond semi-honest to adversarial participants is a potential area of exploration.

\begin{credits}
\subsubsection*{\ackname} This research is part of the Priv-GSyn project, 200021E\_229204 of Swiss National Science Foundation and the DEPMAT project, P20-22 / N21022, of the research programme Perspectief which is partly financed by the Dutch Research Council (NWO).

\subsubsection{\discintname}
The authors have no competing interests to declare that are relevant to the content of this article.
\end{credits}




\bibliographystyle{splncs04}
\bibliography{ref}
\clearpage

\appendix
\section{Datasets and Pre-processing}
\label{dataappendix}
We make use of the following public datasets in our work: Air Quality~\cite{misc_air_quality_360}, SML 2010~\cite{misc_sml2010_274}, PV Power~\cite{pvpower}, Airline Passengers~\cite{airlinepassengers}, and Rossman Sales~\cite{rossmansales}. We do not provide the details on the industrial dataset due to confidentiality agreements. Here we provide a brief description of each dataset and the pre-processing steps applied to the dataset itself. These steps are unconnected with the series pre-processing steps which are part of the \algo framework. For all datasets, we scale all features and output values between the range 0 to 1 using a MinMax scaler to ensure that all datasets have the same range of values for comparing the MSE losses.

\subsection{Air Quality}
 The dataset contains approximately 9300 samples of multivariate time-series data, with 15 attributes. Five of these are true output values on five gases: Carbon Monoxide (CO), Non Metanic Hydrocarbons (NMHC)), Benzene (C6H6)), Total Nitrogen Oxides (NOx), and Nitrogen Dioxide (NO2). Exogenous features such as the temperature, ozone levels, and humidity are provided, along with strongly correlated sensor data for each of the five gases. The dataset contains missing values and duplicates, and contains hourly data for each of the five gases. 

 We preprocess the data by discarding all rows with any missing information and remove duplicate rows. We predict the ground truth values of CO using the other sensor values and information such as temperature and humidity as the exogenous regressors. 

\subsection{SML 2010} 
The SML 2010 dataset contains infomation from a monitor system in a domotic house. It contains approximately 4100 samples with 24 attributes in total, corresponding to 40 days of monitoring data. The attributes contain values such as the indoor and outdoor temperature, lighting levels, Carbon Dioxide levels, relative humidity, rain, windspeed, etc. We predict the indoor habitation temperature using the others as exogenous features.

\subsection{Airline Passengers}
Airline Passengers is a small dataset of 145 samples containing the number of international airline passengers (in thousands) on a monthly basis. The exogenous features are also just two: the year and the month. We predict the number of passengers using the year and month as exogenous regressors.

\subsection{PV Power} The PV Power dataset contains around 3100 samples of solar power generation data from each of two power plants over a 34-day period. Attributes include features such as the DC power, AC power, yield, ambient temperature, irradiation levels, and the data and time. 

We drop identifiers, empty, and duplicate data. We also drop the DC power attribute, and total yield as these features are very strongly correlated with the AC output. As outputs, we predict the AC power generation using the remaining features as exogenous regressors. 

\subsection{Rossman Sales}

The Rossman Sales dataset contains sales data for 1115 store outlets. The attributes consists of features such as holidays, store type, competitor distance, number of customers, and promotional details among others. We predict the sales of the store with ID 1, using the other features as exogenous regressors.

\section{Experimental Evaluation}
\label{experimentappendix}
\begin{figure}[h]
    \centering
    \includegraphics[width=0.75\textwidth]{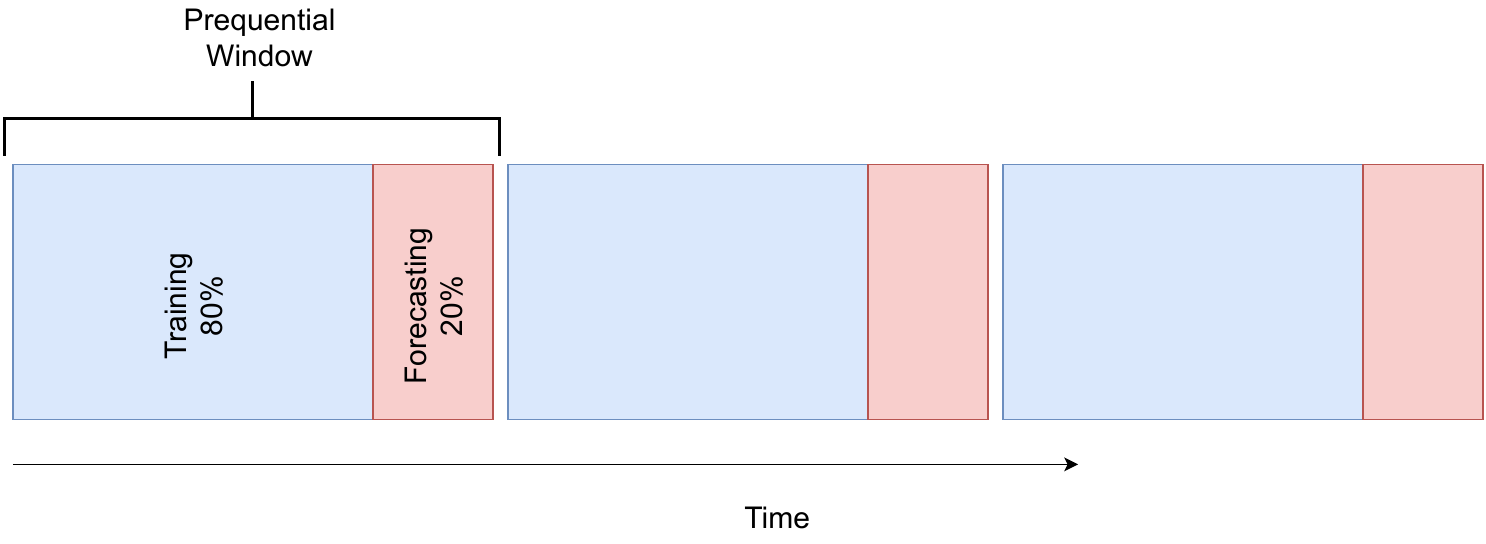}
    \caption{Prequential window evaluation with re-training in every window}
    \label{fig:prequential}
    \vspace{3mm}
\end{figure}

As mentioned in the main text, we use a variation of prequential window testing \cite{cerqueira2020evaluating}, whereby the entire data is broken into windows of a defined length, each one internally split in an 80-20 train-test ratio. This is illustrated in \autoref{fig:prequential}. 

For each window, we train on the portion allotted for training and forecast the remaining. Within a given training window, we first generate the polynomial by processing the time series as in Algorithm 1 from the main text. Identifying the parameters for generating the polynomial can be automated using implementations such as auto arima\footnote{https://alkaline-ml.com/pmdarima/modules/generated/pmdarima.\\arima.auto\_arima.html}.

For each window size, such as 50, 100, 200, 400, we average the MSE loss between the forecasts and the true values across all windows. The average MSE per-window size is given in \autoref{tab:msetabfull}, which is an expanded version of Table 1 from the main text.
\begin{table*}[t]
    \centering
    \begin{tabular}
    {c c c c c c c}
    \Xhline{1.5pt}
    Dataset & Window size & \algos & SARIMAX MLE & LSTM & \algot & $SSSD^{S4}$\\
    \hline
     \multirow{5}{*}{Air quality} &
         50&	0.00071&	0.00110&	0.00244&	0.00198&	0.00270\\
         &100&	0.00059&	0.00048&	0.00055&	0.00075&	0.00086\\
         &200&	0.00066&	0.00124&	0.00069&	0.00067&	0.00100\\
         &400&	0.00080&	0.00068&	0.00087&	0.00061&	0.00144\\
         \hline
         &Avg.&	\textbf{0.00069}&	0.00088&	0.00114&	0.00100&	0.00150\\
         \hline
         \multirow{5}{*}{SML 2010}
         &50&	0.00645&	0.00621&	0.00343&	0.00755&	0.01373\\
         &100&	0.01958&	0.02819&	0.02778&	0.04305&	0.02384\\
         &200&	0.00500&	0.00662&	0.02781&	0.01709&	0.00336\\
         &400&	0.00045&	0.00030&	0.00960&	0.00729&	0.00249\\
         \hline
         &Avg.&	\textbf{0.00787}&	0.01033&	0.01716&	0.01875&    0.01085\\
         \hline
         \multirow{5}{*}{Rossman Sales}
         &50&	0.00070&	0.00079&	0.00464&	0.00183&	0.00589\\
         &100&	0.00064&	0.00071&	0.00340&	0.00496&	0.00281\\
         &200&	0.00068&	0.00086&	0.00674&	0.00088&	0.00233\\
         &400&	0.00095&	0.00072&	0.01079&	0.00206&	0.00220\\
         \hline
         &Avg.&	\textbf{0.00074}&	0.00077&	0.00639&	0.00243&	0.00331\\
         \hline
         \multirow{5}{*}{PV Power}
         &25&	0.00133&	0.00093&	0.00950&	0.00660&	0.00131\\
         &50&	0.00360&	0.00307&	0.01037&	0.00258&	0.00143\\
         &100&	0.00004&	0.00063&	0.01115&	0.00010&	0.00267\\
         &200&	0.00054&	0.00175&	0.54227&	0.00068&	0.00128\\
         \hline
         &Avg.&	\textbf{0.00138}&	0.00159&	0.14333&	0.00249&	0.00167\\
         \hline
         \multirow{6}{*}{Airline Passengers}
         &60&	0.00261&	0.00113&	0.11651&	0.00673&	0.00110\\
         &80&	0.00450&	0.00444&	0.00339&	0.00628&	0.00403\\
         &100&	0.00308&	0.00066&	0.01982&	0.00270&	0.00798\\
         &120&	0.00316&	0.00136&	0.05164&	0.01134&	0.00356\\
         &140&	0.00185&	0.00351&	0.01512&	0.01332&	0.00293\\
         \hline
         &Avg.&	0.00304&	\textbf{0.00222}&	0.04130&	0.00808&	0.00392\\
         \Xhline{1.5pt}
    \end{tabular}
    \caption{Average normalized MSE values for different public datasets, with different prequential window sizes}
    \label{tab:msetabfull}
\end{table*}

\section{SMPC-based XGBoost with VFL}
\label{xgboostappendix}
XGBoost \cite{chen2015xgboost} is a tree-based gradient-boosting algorithm, that iteratively generates an ensemble of trees by greedily learning a new tree at every step to improve on the earlier one. Each tree has weights assigned to its leaf nodes. When making a prediction for a sample, the weights corresponding to the leaf to which the sample was assigned to are summed to give the final prediction score. 

To generate a tree at every iteration, it uses first and second order gradients of the latest predictions, i.e., from the previous tree, in order to set the optimal weights for the new one.

For each sample with index $i$, the first and second order gradients are denoted as $g_i$ and $h_i$, respectively. The sum of the gradients of all instances on a particular node is used to set the new weights for it. For example, for node $j$ and corresponding instance set $I_j$, the accumulated values of $g$ and $h$ are computed as follows:
$G_j = \sum_{i \in I_j}g_i$, and $H_j = \sum_{i \in I_j}h_i$. Based on this, for a tree with $T$ nodes, the weights and objective are calculated as follows:
\begin{equation}
\label{eq:weight}
    w_j = -G_j/(H_j + \lambda) 
\end{equation}
\begin{equation}
\label{eq:obj}
    obj = -0.5 \times \sum_{j=1}^{T}((G_j)^2/(H_j + \lambda)) + \gamma T
\end{equation},
where $\gamma,\lambda$ are regularizers. While eq.\eqref{eq:weight} sets the new weights, eq.\eqref{eq:obj} is used to identify how to split nodes at each iteration. 

With this in mind, using the secret sharing primitives it is possible to compute these functions to extend XGBoost to VFL, which we show in \hyperref[alg:alg6]{Algorithm 6}.

\begin{figure}
    \centering
    \includegraphics[width=0.56\textwidth]{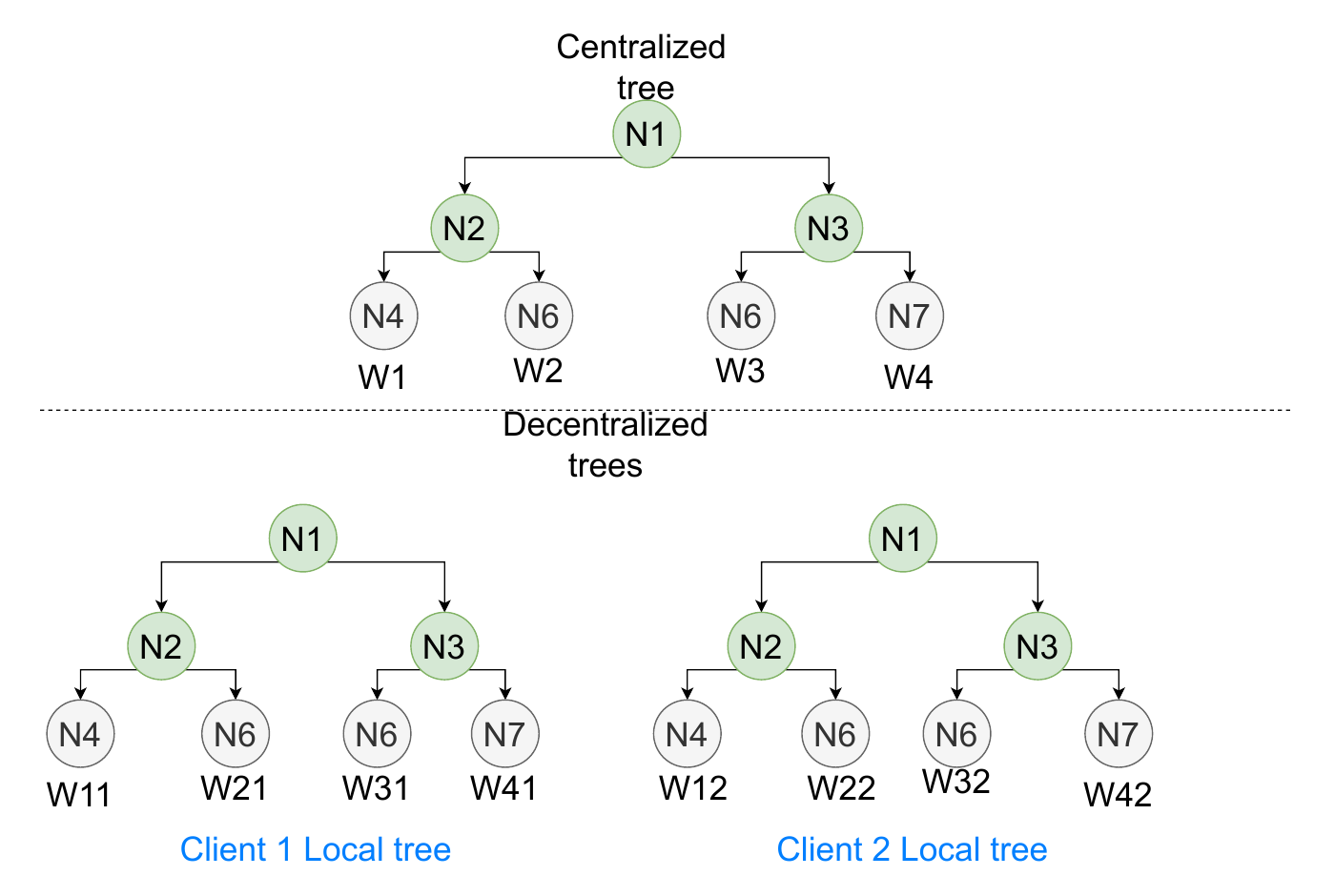}
    \caption{Centralized vs Secretly shared XGBoost trees}
    \label{fig:decentraltrees}
\end{figure}

When XGBoost is trained for VFL using secret sharing, each client obtains a local tree with their own weights as shown in \autoref{fig:decentraltrees}. In the figure, the weights for clients 1 and 2 are distributed such that their local weights are shares of the weights of the centralized version, i.e., $wi = w{i1} + w{i2}$ $\forall i \in [1,4]$

\begin{figure}
    \centering
    \includegraphics[width=0.49\textwidth]{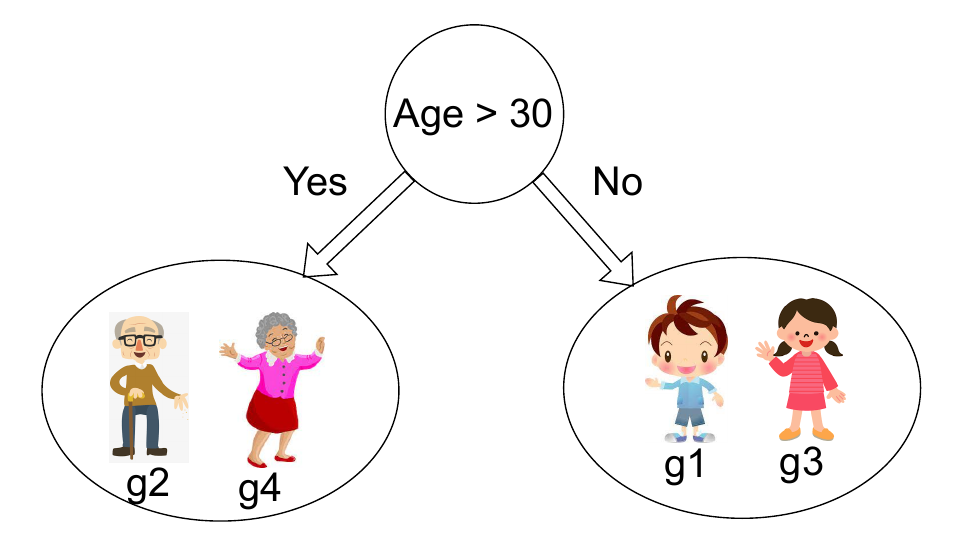}
    \caption{Mapped instances to a node in XGBoost}
    \label{fig:gxgb}
    \vspace{5mm}
\end{figure}

 The indicator vector $s$ on line \ref{ln:indic} of algorithm \ref{alg:alg6}, is a binary vector that is used to point out the location of instances on nodes. We explain this with the help of the example in \autoref{fig:gxgb}. To calculate the updated weight of the node with instances  that have an age greater than 30 (bottom left), we need to find the sum of $\sum_{i \in I_j}g_i$, where $I_j = \{2,4\}$ eq.\eqref{eq:weight}. The indicator vector in this case would be $s = [0,1,0,1]$, meaning that nodes 2 and 4 are part of node $j$. If we have a vector of the gradients, $g=[g1, g2, g3, g4]$, we can compute $g_2+g_4$ as $s \odot g$, i.e., the inner product. Under VFL, both the gradients $g$, and the indicator vector $s$, exist as secret shares across clients, i.e., $s = \sum_{k}^{K}$\SHR[k]{s_k}, and $g = \sum_{k}^{K}$\SHR[k]{g_k}. Therefore, we can compute the product using MPC primitives for matrix multiplication.

The process of split-finding and setting weights is done using the \bemph{SecureBuild} function, which makes use of secret sharing primitives to compute the functions in eq.\eqref{eq:obj} and eq.\eqref{eq:weight}. We defer readers to the implementation in Xie et al.~\cite{xie2022efficient} for additional details on this.

\begin{algorithm}[tb]
    \caption{Fit XGBoost \protect\cite{xie2022efficient}}
    \label{alg:alg6}
    \textbf{Data}: Secretly shared matrices: \SHR{X}, \SHR{Y}, \SHR{\hat{Y}} across $K$ clients, $C_1,C_2,..,C_K$\\
    \textbf{Output}: Learned tree for the current iteration, one on each client.
    \begin{algorithmic}[1] 
    \STATE Initialize indicator vector $s \gets 1$ on all $C_k$ \label{ln:indic}
    \IF {active client $C_1$}
        \STATE Compute derivatives $g,h$
        \STATE Generate shares \SHR{g}, \SHR{h}, \SHR{s}
    \ENDIF
   \STATE $Tree_k = \bemph{SecureBuild}$(\SHR[k]{g},\SHR[k]{h}, \SHR[k]{s}) on each $C_k$
    \RETURN $Tree_k$ on $C_k$
    \end{algorithmic}
\end{algorithm}

\end{document}